(RESEARCH ARTICLE)

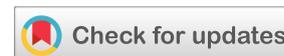

# Features extraction for image identification using computer vision

Venant Niyonkuru [1, *], Sylla Sekou [2] and Jimmy Jackson Sinzinkayo [3]

[1] Department of Computing and Information System, Kenyatta University, Kenya.
[2] Department of Mathematics, Institute for Basic Science, Technology and Innovation, Pan-African University, Kenya.
[3] Department of Software Engineering, College of Software, Nankai University, China.

## Abstract

This study examines various feature extraction techniques in computer vision, the primary focus of which is on Vision Transformers (ViTs) and other approaches such as Generative Adversarial Networks (GANs), deep feature models, traditional approaches (SIFT, SURF, ORB), and non-contrastive and contrastive feature models. Emphasizing ViTs, the report summarizes their architecture, including patch embedding, positional encoding, and multi-head self-attention mechanisms with which they overperform conventional convolutional neural networks (CNNs). Experimental results determine the merits and limitations of both methods and their utilitarian applications in advancing computer vision.

## 1. Introduction

Feature extraction is a critical stage in the computer vision domain that is the backbone of transforming raw image data with high amounts into compact, descriptive representations that enable object detection, image categorization, segmentation, and scene interpretation. Traditionally, feature extraction methods have developed over time based on the need for creating descriptors that are invariant to scaling, rotation, lighting, and perspective, but computationally effective (Dosovitskiy et al, 2020; Jiang, 2009; Lowe,2004; Grill et al, 2020).

Traditional feature extraction techniques like Scale-Invariant Feature Transform (SIFT), Speeded-Up Robust Features (SURF), and Oriented FAST and Rotated BRIEF (ORB) have been instrumental for initial computer vision systems (Lowe,2004; Rublee et al, 2011, Morrow, 2000) . These algorithms engineer features from local image properties, finding keypoints and constructing descriptors to facilitate matching among different images. While resistant in the majority of scenarios, these hand-crafted features are often prone to difficulty with complexity, scalability, and sometimes devoid of semantic context.

Deep learning transformed feature extraction by the power to learn hierarchical representations directly from data without needing hand-designed features. Convolutional Neural Networks (CNNs) emerged as the standard by leveraging local spatial correlation and shared weights but with the expense of local receptive fields, which limit their capacity to learn long-range dependencies in images (Dosovitskiy et al, 2020; Ali, et al, 2023; Krizhevsky et al, 2012,Morrow, 2000). Here, Vision Transformers (ViTs) have emerged as a highly promising substitute that brings the self-attention mechanism of NLP into computer vision (Dosovitskiy et al, 2020). ViTs work by dividing images into fixed-size patches, flattening them, and linearly embedding them. Positional embeddings help to maintain the spatial information, and the patch embedding is fed into multi-head self-attention to capture global context (Dosovitskiy et al, 2020, Patwardhan et al, 2023, Montrezol, 2024). This paradigm change helps ViTs capture the relationships of the entire image and surpass limitations intrinsic to CNNs and delivering superior performance across a wide array of vision benchmarks.

[*] Corresponding author: Venant Niyonkuru

Also, newer architectures such as Generative Adversarial Network (GAN)-based models and contrastive learning techniques have added to the list of tools used to learn semantic features from images (Ali et al, 2024, Cao et al, 2018; Kovács et al, 2023).

These are aimed at learning discriminative and generative representations that are useful over a broad range of tasks ranging from image generation to self-supervised learning (Grill et al., 2020; Ansar et al., 2024). This study comprehensively examines these varied feature extraction approaches, demystifying the principle behind Vision Transformers and their position within the wider computer vision context.

Experimental results clarify their individual strengths, compromises, and practical usability, sketching the outline for the best feature extraction approaches to use in real-world applications(Purchase, 2012).

## 2. Related work

Traditional approaches are SIFT (Lowe, 2004), SURF (Bay et al., 2008), and ORB (Rublee et al., 2011), which have served as standard baseline approaches to image matching and recognition. These approaches rely on handcrafted descriptors to obtain local features. They operate well in structured or low-variation visual scenes. However, they cannot deal with scale variation, illumination variation, and occlusion. One of the major breakthroughs as exemplified by the emergence of deep learning models, particularly Convolutional Neural Networks (CNNs), was when these models learned to learn end-to-end discriminative hierarchical features from raw images (Krizhevsky et al., 2012). CNNs were more generalizable on a wide variety of vision tasks and thus remained the standard for a number of years.

In recent times, Vision Transformers (ViTs) have been strong competitors that are based on self-attention mechanisms for obtaining long-range relations in images (Dosovitskiy et al., 2020). ViTs outperformed CNNs on big-benchmark benchmarks, particularly when they were trained on very large datasets. Simultaneously, Generative Adversarial Networks (GANs) have not only been utilized for image synthesis but also for feature extraction, depending on discriminators for obtaining detailed, high-level features. Furthermore, contrastive learning techniques such as BYOL (Grill et al., 2020) and SimCLR have enhanced self-supervised feature learning by optimizing the agreement between multiple copies of an image that are transformed differently.

Recent large-scale surveys (Ali et al., 2023; Patwardhan et al., 2023) cover developments in these architectures, presenting trends and open questions. However, there are fewer papers providing an explicit comparison of these different approaches under the same experimental setting. This paper fills this gap by comparing classical descriptors, CNNs, ViTs, and GAN-based models on an identical setup of popular benchmarks and measures.

## 3. Methodology

### 3.1. Vision Transformer (Vits)

*3.1.1. Definition and functionality*

Vision Transformers (ViTs) are deep learning models that leverage self-attention mechanisms to process image data, offering improved performance over traditional convolutional neural networks (CNNs) (Dosovitskiy et al, 2020) .

*3.1.2. Architecture*

ViTs divide an image into fixed-size patches, linearly embed them, and feed them into a transformer encoder. The key components include:

- Patch Embedding Layer: Converts image patches into token embeddings.
- Positional Encoding: Adds spatial information to tokens.
- Multi-Head Self-Attention: Captures long-range dependencies in an image.
- Feed-Forward Network (FFN): Processes token representations for classification tasks.

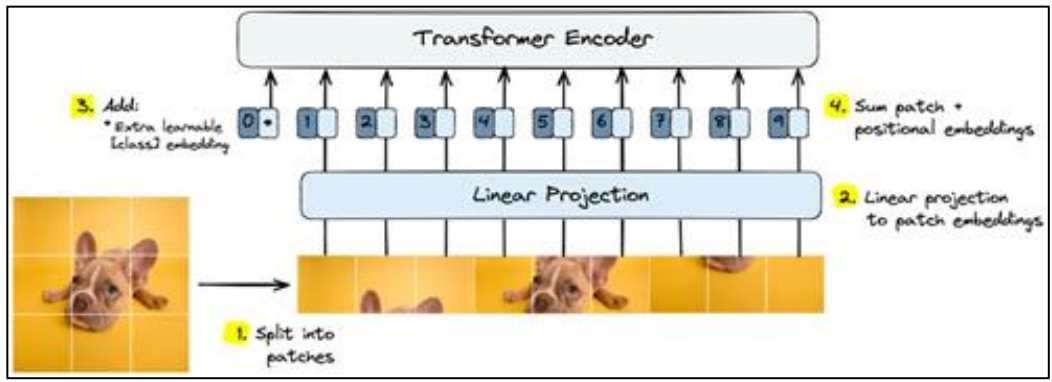

**Figure 1** Transformer Encorder

How ViTs Work

- The image is broken into non-overlapping patches
- Each patch is flattened and subsequently passed through a linear projection.
- The transformer encoder converts the patch embeddings through self-attention.
- classification head produces predictions from the last encoded representation.

*3.1.3. Image Patches*

The process starts with dividing an image into small, fixed-size patches, and that is a simple transformation step. This process has a direct analogy in natural language processing (NLP) where a sentence is segmented into individual units such as words or subword tokens. Just like how every token within a sentence carries contextual meaning, every patch within an image captures localized visual context. In this analogy, the entire image is taken as a sentence, and its patches are akin to tokens, which enable transformer-based models originally designed for text to be used on visual data.

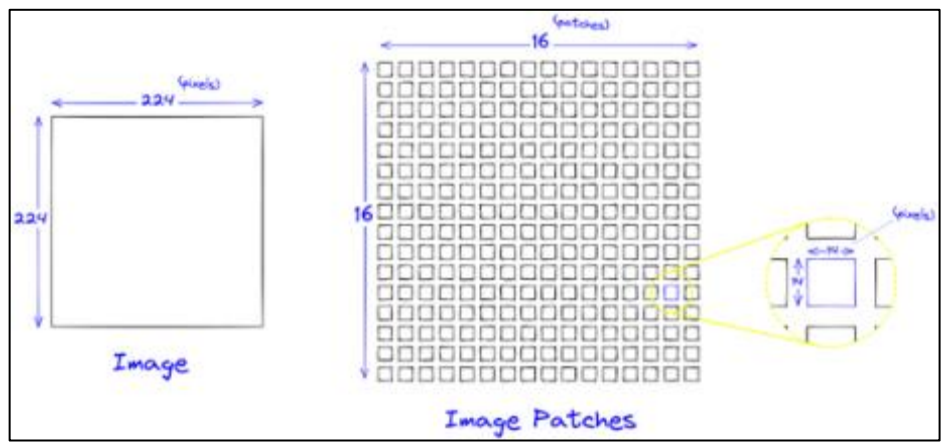

**Figure 2** Image to Image Patches

Both vision transformers (ViT) and natural language processing (NLP) partition large inputs (i.e., sentences in text or entire images into smaller ones, e.g., tokens in text or image patches). For instance, processing an entire 224×224 pixel image directly would entail an impossibly large number of calculations, approximately 2.5 billion comparisons. But by dividing the very same image into 256 patches, each 14×14 pixels, the computation load of one attention layer becomes incredibly smaller approximately 9.8 million comparisons.

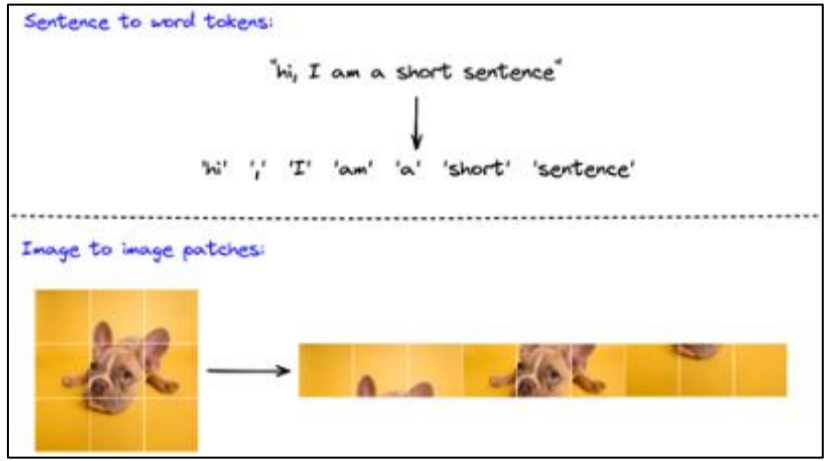

**Figure 3** Vision transformers

### 3.1.4. Linear Projection

Following patch division of the image, each patch is then converted from a 2D array to a 1D vector using a linear projection, effectively projecting raw pixel information into a set of patch embeddings.

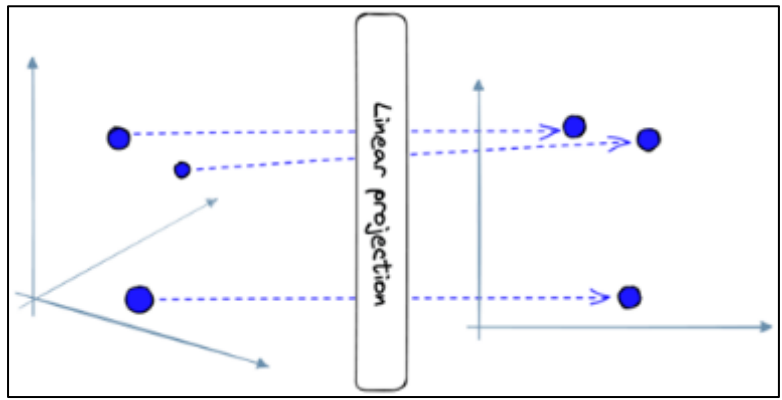

**Figure 4** Linear Projection

The role of the linear projection layer is to transform each image patch into a fixed-size vector representation, the aim being to maintain meaningful relations so visually similar patches produce similar embeddings. This transformation brings the data into a form compatible with the input format needed by the transformer model. Two further processing steps remain before these embeddings can be used.

### 3.1.5. Learnable Embeddings

One of the important features added in widely used transformer models such as BERT is the inclusion of a special classification token, also known as [CLS]. This token is placed at the beginning of every input sequence and is meant to capture the sentence-level representation for classification tasks.

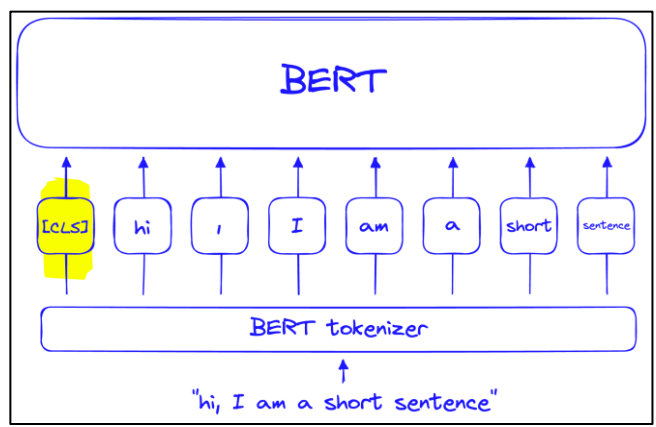

**Figure 5** Bert Tokenizer

There is a unique token, [CLS], in BERT that is added to the beginning of all input sequences. This token is embedded like any other and passed through the encoder layers of the model. The [CLS] token is special in that it doesn't represent any specific word of the input it begins as a neutral or uninitialized vector. In addition, during pretraining, this final output at the [CLS] position is fed as input to a classification layer. This encourages the model to encode information from the entire sentence into this single vector, learning an effective representation of the input. Vision Transformers (ViT) do exactly the same thing with a learnable embedding that serves the same purpose as the [CLS] token in BERT, providing a summary representation for image-level classification tasks.

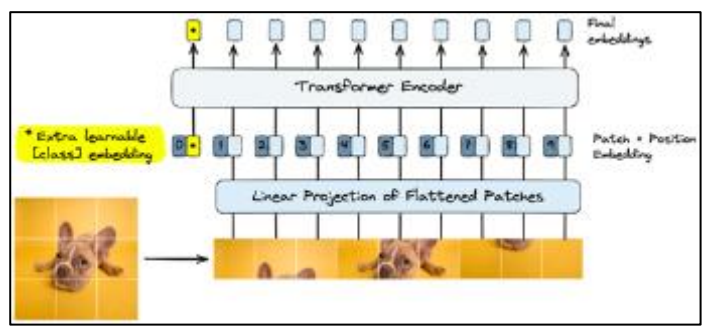

**Figure 6** Transformer Encoder to Linear Projection

*3.1.6. Positioning Embedding*

Transformers do not have an inherent perception of sequence or spatial arrangement of input tokens or patches. However, preserving order is important in language, where word reordering can dramatically alter meaning. The same is true for visual information: when the components of a picture are mixed up, as in a jigsaw puzzle, identification of the whole picture becomes extremely challenging. This is also true for transformer models, which require an additional mechanism to understand the relative position of these parts.

To address this, positional embeddings are added. In Vision Transformers (ViT), these are learned and of the same dimension as the patch embeddings. Following the division of the image into patches and adding the special classification token, each element is added to its respective positional embedding. These position vectors are also trained along with the model and can further be fine-tuned later. They gradually come to denote spatial relationships, usually identical to proximate locations in the grid particularly in the same column or row such as:

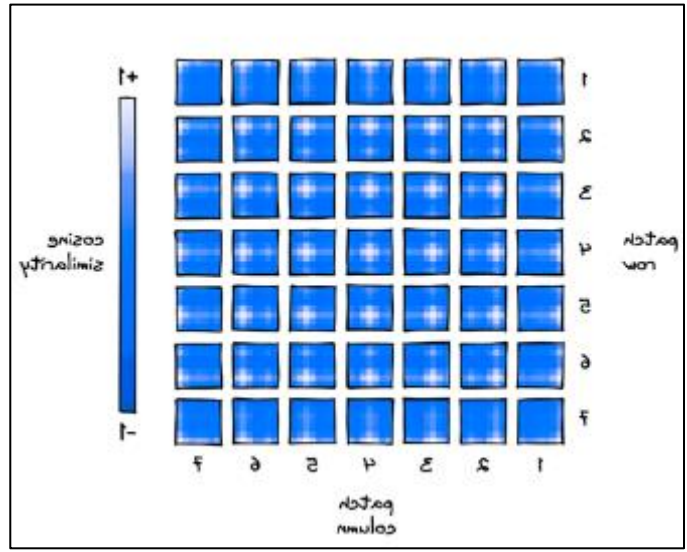

**Figure 7** Embeddings Position

Once positional embeddings are added, the patch embeddings are complete. These enhanced embeddings are then passed into the Vision Transformer (ViT), and they are processed in the same way as regular tokens in a standard transformer model

Imprementation

```
# import CIFAR-10 dataset from HuggingFace
from datasets import load_dataset
dataset_train = load_dataset(
    'cifar10',
    split='train', # training dataset
    ignore_verifications=False  # set to True if seeing splits Error
)
dataset_train
Out[2]:
Dataset({
    features: ['img', 'label'],
    num_rows: 50000
})
dataset_test = load_dataset(
    'cifar10',
    split='test', # training dataset
    ignore_verifications=True  # set to True if seeing splits Error
)
dataset_test
Out[3]:
Dataset({
    features: ['img', 'label'],
    num_rows: 10000
})
```

The training dataset consists of 60,000 images across 11 unique classes. In order to obtain the equivalent human-readable labels for these classes, the following steps may be used:

```
# check how many labels/number of classes
num_classes = len(set(dataset_train['label']))
labels = dataset_train.features['label']
sses, labels

Out[4]:
(11
```

ClassLabel has 11 classes: ['airplane', 'automobile', 'bird', 'cat', 'deer', 'dog',..].

Each entry in the dataset contains two features: `img` and `label`. The `img` feature contains a 32x32 pixel image which is of type PIL and with three color channels of RGB (red, green, blue).

```
dataset_train[0]
Out[5]:
{'img': <PIL.PngImagePlugin.PngImageFile image mode=RGB size=32x32 at 0x1477E4880>,
 'label': 0}
dataset_train[0]['img']
<PIL.PngImagePlugin.PngImageFile image mode=RGB size=32x32 at 0x16B7658E0>

dataset_train[0]['label'], labels.names[dataset_train[0]['label']]
Out[7]:
(0, 'airplane')
```

*3.1.7. Feature extraction*

Before sending images to the Vision Transformer (ViT) model, a feature extractor is used to handle preprocessing. This involves resizing and normalizing images, converting them into tensors referred to as "pixel_values."

```python
from transformers import ViTFeatureExtractor
# import model
model_id = 'google/vit-base-patch16-224-in21k'
feature_extractor = ViTFeatureExtractor.from_pretrained(
    model_id
)
feature_extractor
```

```
Out[9]:
ViTFeatureExtractor {
  "do_normalize": true,
  "do_resize": true,
  "feature_extractor_type": "ViTFeatureExtractor",
  "image_mean": [
    0.5,
    0.5,
    0.5
  ],
  "image_std": [
    0.5,
    0.5,
    0.5
  ],
  "resample": 2,
  "size": 224
}
```

The feature extractor may be initialized with the Transformers library of Hugging Face, as shown below:

The output size is set by "size" at 224x224 pixels.
To process an image with the feature extractor, we do the following:

```python
example = feature_extractor(
    dataset_train[0]['img'],
    return_tensors='pt'
)
```

The feature extractor configuration shows that normalization and resizing are set to true. Normalization is performed across the three color channels using the mean and standard deviation values stored in "image_mean" and "image_std" respectively.

Therefore, it is optimal to use an image that is slightly larger than needed, since reducing by a small amount usually preserves visual quality and avoids introducing visible degradation in image quality.

```
Out[10]:
{'pixel_values': tensor([[[[ 0.3961,  0.3961,  0.3961,  ...,  0.2941,  0.2941,  0.2941],
          [ 0.3961,  0.3961,  0.3961,  ...,  0.2941,  0.2941,  0.2941],
          [ 0.3961,  0.3961,  0.3961,  ...,  0.2941,  0.2941,  0.2941],
          ...,
          [-0.1922, -0.1922, -0.1922,  ..., -0.2863, -0.2863, -0.2863],
          [-0.1922, -0.1922, -0.1922,  ..., -0.2863, -0.2863, -0.2863],
          [-0.1922, -0.1922, -0.1922,  ..., -0.2863, -0.2863, -0.2863]],

         [[ 0.4824,  0.4824,  0.4824,  ...,  0.3647,  0.3647,  0.3647],
          [ 0.4824,  0.4824,  0.4824,  ...,  0.3647,  0.3647,  0.3647],
          [ 0.4824,  0.4824,  0.4824,  ...,  0.3647,  0.3647,  0.3647],
          ...,
          [-0.2784, -0.2784, -0.2784,  ..., -0.3961, -0.3961, -0.3961],
          [-0.2784, -0.2784, -0.2784,  ..., -0.3961, -0.3961, -0.3961],
          [-0.2784, -0.2784, -0.2784,  ..., -0.3961, -0.3961, -0.3961]]]])}
example['pixel_values'].shape
Out[11]:
torch.Size([1, 3, 224, 224])
```

Evaluation and Prediction

The Trainer evaluates during training but we can also quickly do a more qualitative verification (or estimation) by passing through a single image with the model and feature_extractor.

We will pass the following image:

```
# show the first image of the testing dataset
image = dataset_test["img"][0].resize((200,200))
Image
Out[50]:
<PIL.Image.Image image mode=RGB size=200x200 at 0x7FA9D072E0A0>
```

The picture is of poor visual quality and does not have distinguishing features, so visual categorization based on the picture is difficult.

However, the label given classifies the subject as a cat. We will now go ahead and test the model's prediction for this picture.

```
# Import fine-tuned version of model from Hugging Face hub (if necessary)
model_id = 'LaCarnevali/vit-cifar10'
model = ViTForImageClassification.from_pretrained(model_id)
inputs = feature_extractor(image, return_tensors="pt")
with torch.no_grad():
    logits = model(**inputs).logits
predicted_label = logits.argmax(-1).item()
labels = dataset_test.features['label']
labels.names[predicted_label]
Out[61]: 'cat'
```

Seems like the model is correct!

That's it for our tour of the Vision Transformer and how to use it with Hugging Face Transformers. It's truly remarkable how quickly transformers have taken over natural language processing and are now making significant incursions into computer vision.

Just a few years back, before 2021, it was inconceivable to use transformers anywhere except in NLP. But despite being branded as "those language models," transformers are today at the core of some of the world's most state-of-the-art computer vision systems. They're components of next-generation architectures like diffusion models [6, 12, 15], and even power aspects of Tesla's Full Self Driving feature [14, 13, 15, 6].

As time proceeds, we can anticipate a greater convergence of NLP and vision, with transformers persisting in leading the way of innovation across both fields.

## 4. Conclusion

To put it briefly, the explosive growth of feature extraction algorithms is an enormous step forward for computer vision system competence. Handcrafted solutions such as SIFT, SURF, and ORB set the groundwork by providing stable and interpretable descriptors that are extremely good in specific instances. However, they do poorly when they are tested against complex, large-scale, or high-variant visual information.

The introduction of deep learning techniques in the form of Vision Transformers (ViTs) is a revolution in paradigm shifting. By adopting the self-attention mechanism in examining the picture holistically, ViTs have the ability to take advantage of intricate relationships tricky to model in isolation using convolutional methods alone. Their ability to include spatial information and condition image patches alike like tokens from NLP have discovered new potential for better recognition, classification, and downstream visual tasks with high accuracy and robustness.

Furthermore, the inclusion of transformers in autonomous cars such as Tesla's Full Self Driving, and diffusion models, showcases the growing real-world practical applicability and adaptability of self-attention-based architectures. The convergence of ideas in NLP and computer vision holds the potential for a future where multimodal models could perhaps be easily able to handle all kinds of data.

Though their power is remarkable, Vision Transformers are not without problems, including their extensive training data needs and computational costs. As research continues, hybrid architectures, training methods, and streamlined transformer variants are being developed to counter these limitations.

In summary, recognizing the advantages and limitations of different feature extraction techniques ranging from classical to transformer models is vital in pushing computer vision further. As transformers progressively gain the stage, drawing inspirations from across fields, the next few years promise to be huge in the realization of leaps that will continue to close the gaps between perception, reasoning, and cognition in artificial systems.

## Compliance with ethical standards

*Disclosure of conflict of interest*

No conflict of interest to be disclosed.